%% file: acl_latex.tex
\documentclass[11pt]{article}

\usepackage[final]{acl}

\usepackage{times}
\usepackage{latexsym}
\usepackage{amsmath}

\usepackage[T1]{fontenc}

\usepackage[utf8]{inputenc}

\usepackage{microtype}

\usepackage{inconsolata}

\usepackage{graphicx}
\usepackage[size=tiny]{todonotes}
\usepackage{booktabs}   
\usepackage{makecell}   
\usepackage{adjustbox}  
\usepackage{graphicx}   
\usepackage{tcolorbox}
\usepackage{enumitem}    
\usepackage{placeins} 
\usepackage[table]{xcolor} 
\def\checkmark{\tikz\fill[scale=0.4](0,.35) -- (.25,0) -- (1,.7) -- (.25,.15) -- cycle;} 
\definecolor{darkgreen}{rgb}{0.0, 0.5, 0.0}
\definecolor{rowgray}{gray}{0.95}
\usepackage{float}
\usepackage{subcaption}
\usepackage{tabularx}
\usepackage{array}

\title{%
  \makebox[\textwidth][c]{
    \raisebox{-0.42\height}{\includegraphics[width=2.2em]{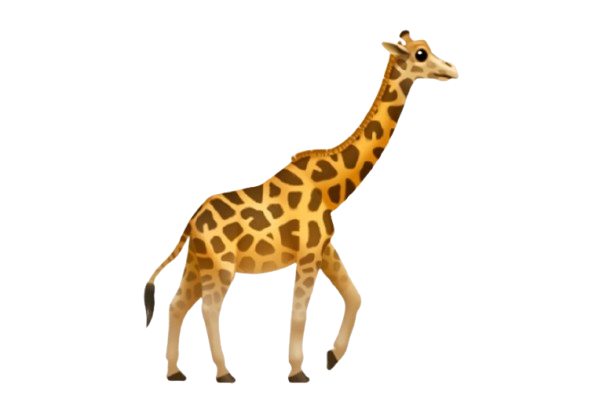}}
    \hspace{0.6em}%
    \parbox[t]{0.67\textwidth}{\centering\bfseries GerAV: Towards New Heights in German Authorship Verification using Fine-Tuned LLMs on a New Benchmark}
  }
}

\author{
Lotta Kiefer\textsuperscript{1}\thanks{Equal contribution},
Christoph Leiter\textsuperscript{2}\footnotemark[1],
Sotaro Takeshita\textsuperscript{3},
Elena Schmidt,\\
\textbf{Steffen Eger\textsuperscript{4}}\\
\textsuperscript{1,2,3,4}University of Technology Nuremberg (UTN) \\
{\small \textsuperscript{1}lotta.kiefer@utn.de \;
\textsuperscript{2}christoph.leiter@utn.de \;
\textsuperscript{4}steffen.eger@utn.de}
}

\newcommand{\shortname}{GerAV}
\newcommand{\dataset}{GerDATA}

\newcommand{\green}[1]{\textcolor{darkgreen}{#1}}

\begin{document}
\maketitle
\begin{abstract}
Authorship verification (AV) is the task of determining whether two texts were written by the same author and has been studied extensively, predominantly for English data. In contrast, large-scale benchmarks and systematic evaluations for other languages remain scarce. We address this gap by introducing GerAV, a comprehensive benchmark for German AV comprising over 400k labeled text pairs. GerAV is built from Twitter and Reddit data, with the Reddit part further divided into in-domain and cross-domain message-based subsets, as well as a profile-based subset. This design enables controlled analysis of the effects of data source, topical domain, and text length. Using the provided training splits, we conduct a systematic evaluation of strong baselines and state-of-the-art models and find that our best approach, a fine-tuned large language model, outperforms recent baselines by up to 0.09 absolute F1 score and surpasses GPT-5 in a zero-shot setting by 0.08. We further observe a trade-off between specialization and generalization: models trained on specific data types perform best under matching conditions but generalize less well across data regimes, a limitation that can be mitigated by combining training sources. Overall, GerAV provides a challenging and versatile benchmark for advancing research on German and cross-domain AV.\footnote{Our code and information about data access are available on \href{https://github.com/NL2G/GerAV/}{GitHub}.}
\end{abstract}

\input{text/1_introduction}

\input{text/2_related}

\input{text/3_data_curation}
\input{text/4_Experiment_Setup}
\input{text/5_results}

\input{text/7_conclusion}
\clearpage

\input{text/8_limitations}
\input{text/9_ethics}
\input{text/10_acknowledgements}
\bibliography{custom}

\appendix

\input{text/11_appendix}

\end{document}

%% file: text/1_introduction.tex
\section{Introduction}
\label{sec:introduction}

\begin{figure}[!t]
    \centering
    \includegraphics[width=\linewidth]{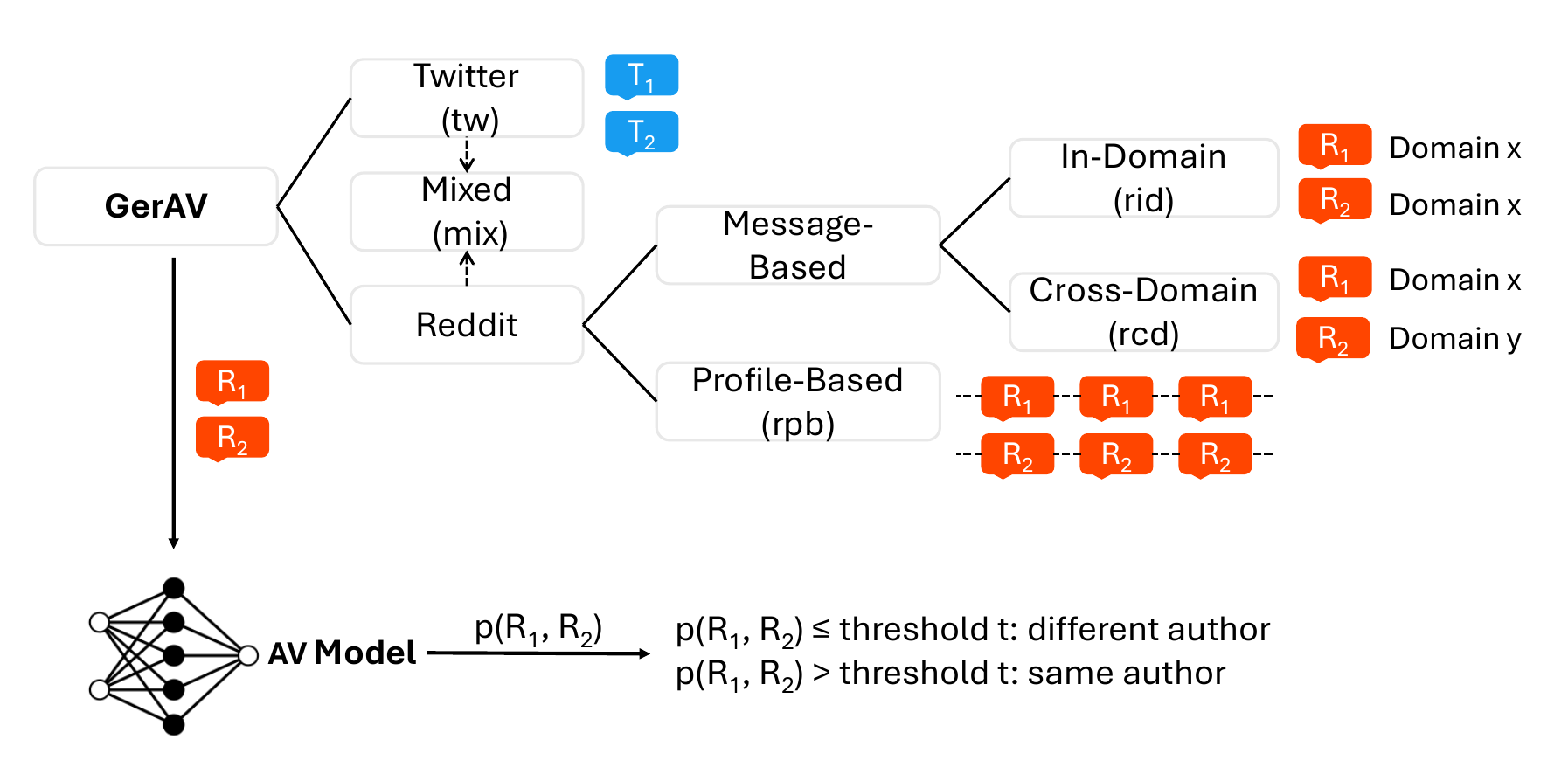}
    \caption{Overview of our approach. The figure presents the GerAV benchmark and its structure, illustrating how the datasets are used as input for AV models during training and testing. Based on a decision threshold, the models predict whether a given pair of texts was written by the same or by different authors. Each pair consists of one or several posts from a single author, contrasted with post(s) from either the same or another author. The figure represents these post pairs as $T_1$ and $T_2$ (Twitter, in blue) and $R_1$ and $R_2$ (Reddit, in orange).}
    \label{fig:gerav_overview}
\end{figure}

Authorship analysis aims to assess the likelihood that a text was produced by a particular individual based on stylistic characteristics. 
It has important applications in a range of domains, including plagiarism detection, the analysis of misinformation spread, and forensic investigations \cite{ramnath2025cave}. In particular, the German Federal Criminal Police Office (Bundeskriminalamt) reports the use of author recognition as a forensic tool when written texts are relevant to criminal offenses \cite{bka2025autorenerkennung}. Given the sensitive and high-stakes nature of these applications, the development of reliable and high-performing methods for authorship analysis is of critical importance.

Authorship analysis is typically divided into two closely related tasks: authorship attribution (AA) and authorship verification (AV). While AA identifies the author of a text from a closed set of candidates, AV determines whether two given texts were written by the same author. This work focuses on AV, which better reflects real-world forensic scenarios where candidate authors are unknown.

While authorship analysis has received considerable attention in the Natural Language Processing (NLP) literature, particularly through feature-based and neural approaches \cite{tyo2023valla}, the use of large language models (LLMs) remains underexplored \cite{huang2025authorship}. Moreover, most existing methods are evaluated primarily on English benchmarks, a limitation that is especially problematic given that authorship analysis relies on language-specific stylistic and linguistic dependencies.

To address these gaps, we make the following contributions: 
\begin{itemize}
  \item[\green{\checkmark}] \textbf{\textit{GerAV} benchmark}: We introduce GerAV, one of the largest benchmarks for the evaluation of \textbf{Ger}man \textbf{A}uthorship \textbf{V}erification. Comprising a substantial number of German-language online posts (over 400k message pairs), it enables robust and diverse evaluation scenarios. The benchmark includes subsets from different data sources and domains, allowing us to analyze their effects on the evaluation models. Additionally, we provide a profile-based subset in which messages from each author's profile are concatenated to evaluate model performance when more data is available. 
  \item[\green{\checkmark}] \textbf{Broad comparison of baseline models}: We perform a broad evaluation of established AV methods, recent approaches, and zero-shot LLMs, which, to our knowledge, is the most comprehensive AV evaluation to date that includes recent LLMs.
  \item[\green{\checkmark}] \textbf{State-of-the-art models for German AV}: Using distinct train splits of the GerAV datasets, we fine-tune several open-source LLMs for the task of German AV. By doing so, several model versions strongly outperform all existing baselines. Most notably, Gemma-3-12b, tuned on a mix of train splits from all GerAV subsets, achieves an average F1-Score of 0.83 across all test splits. This is 0.09 higher than the best baseline at 0.74. On a small test subset, this model also beats GPT-5 by 0.08 points F1-Score while being significantly smaller in size.
  \item[\green{\checkmark}] \textbf{Exploring cross-domain and message length}: We explore performance differences when tuning and evaluating on in-domain vs.\ cross-domain data and on single messages vs.\ whole profiles. The baseline methods' performance (including LLMs) increases with input length, from an F1-score of approx.\ 0.78 at 25 words to up to almost 1.0 at 900 words. For our tuned models, this is only the case if the training data contains long examples. Also, the average performance across all models on the GerAV cross-domain subset is 0.06 points lower than on the in-domain subset. Tuning on cross-domain data solves this issue, but leads to weaker performance on in-domain data and other datasets.
 \item[\green{\checkmark}] \textbf{Investigating low-resource and cross-language scenarios}: We evaluate the robustness of our approach under reduced training data conditions and find that decreasing the number of training samples to one quarter results in only a 0.04 drop in F1. Furthermore, we highlight the importance of language-specific AV benchmarks and models by applying a model trained on English Twitter data to our GerAV Twitter test set, where we observe a strong decrease of 0.12 in F1.
\end{itemize}
Figure \ref{fig:gerav_overview} displays an overview of our approach.

%% file: text/2_related.tex
\section{Related Work}
\label{sec:related_work}

\paragraph{Feature-Based and Neural Approaches} 
In the NLP literature, AA and AV methods can be broadly categorized into three classes: feature-based approaches, neural models, and methods leveraging LLMs. Feature-based methods rely on handcrafted representations derived from morphological, syntactic, and semantic properties, such as character n-gram frequencies, part-of-speech variability, and semantic dependency features \cite{stamatatos2009survey}, with short character n-grams being especially effective \cite{grieve2007quantitative, stamatatos2006ensemble}. Although these approaches are highly interpretable and offer transparent explanations, they typically underperform compared to more complex models \cite{zeng2025residualized}.

Neural approaches alleviate the need for explicit feature engineering by learning representations directly from data. Prior work has explored recurrent neural networks (RNNs) \cite{gupta2019authorship}, long short-term memory networks (LSTMs) \cite{qian2017deep}, convolutional neural networks (CNNs) \cite{hossain2021authorship}, and attention-based siamese architectures \cite{boenninghoff2019explainable}. More recently, pretrained transformer models such as BERT \cite{devlin2019bert}, its variants, and Sentence-BERT \cite{reimers-gurevych-2019-sentence} have achieved strong performance in AA and AV \cite{fabien2020bertaa, manolache2021transferring, rivera2021learning, schlicht2021unified}. While these models offer superior performance and domain adaptability, they generally lack the interpretability of feature-based methods.

\paragraph{Large Language Models} 
Authorship analysis, like many other NLP tasks, is increasingly influenced by advances in LLMs. \citet{huang2024can} investigate zero- and few-shot prompting strategies with GPT-3.5 \cite{openai_gpt-3.5} and GPT-4 \cite{openai_gpt-4}, showing that linguistically informed prompting (LIP) can outperform traditional baselines without task-specific training. However, reliance on online APIs raises concerns about reliability, reproducibility, and data privacy \cite{ramnath2025cave}, limiting use in sensitive domains such as forensic analysis. To overcome this, \citet{ramnath2025cave} employ offline LLMs in zero- and few-shot settings and further improve performance through fine-tuning. Generating rationales through carefully designed prompts enhances explainability, reduces output variability, and surpasses both GPT-4 zero-shot and feature-based baselines in their work. Similarly, \citet{hu2024instructav} explore instruction-tuning for AV using OPT \cite{zhang2022opt} and LLaMA \cite{touvron2023llama2openfoundation, touvron2023llamaopenefficientfoundation} models, integrating explanations directly into the model output and benchmarking against BERT-based and zero-shot approaches.

Despite these advances, the potential of LLMs for authorship analysis remains underexplored \cite{huang2025authorship}. Existing studies are limited in scope, often evaluating only a small number of models or baselines. For example, \citet{ramnath2025cave} and \citet{hu2024instructav} each compare their approaches against only a single non-LLM baseline method, which is insufficient to establish superiority over alternative baselines that have demonstrated strong performance in prior work. Although several surveys and shared tasks exist for AA and AV \cite{stamatatos2009survey, kestemont2020overview, bevendorff2021overview, he2024authorship, tyo2023valla}, LLMs are largely absent from these evaluations. Moreover, comparisons across studies are hindered by inconsistent datasets, splits, and evaluation metrics \cite{tyo2023valla}, underscoring the need for systematic and comprehensive benchmarking including LLMs. We address this gap by conducting a detailed evaluation of zero-shot and tuned LLMs against a wide range of high-performing baselines.

\paragraph{Multilingual Settings}
Authorial style is inherently shaped by language-specific grammar, morphology, and writing conventions, making language-aligned evaluation essential. Without it, models risk capturing language-dependent artifacts rather than true stylistic signals. While some work has explored multilingual modeling \cite{qiu-etal-2025-mstyledistance, kim-etal-2025-leveraging}, non-English resources remain limited. Few datasets address this gap: \citet{israeli2025million} introduce the One Million Author Corpus with Wikipedia articles across dozens of languages, including over 100,000 German authors, with cross-domain samples from Wikipedia namespaces; \citet{murauer2019generating} construct a multilingual AV dataset from Reddit spanning many languages, including ~80,000 German posts, using subreddit membership to induce cross-topic variation; and \citet{halvani2016authorship} evaluate AV across five languages, including a small German corpus from news, reviews, forums, and novels. Closest to our work, \citet{boenninghoff-etal-2024-wrote} present a German AV dataset derived from a newspaper discussion forum. However, their work focuses on author diarization in chronologically ordered text streams, whereas our benchmark emphasizes diverse and realistic evaluation settings tailored to forensic applications.

This work contributes to language-specific AV research by introducing a comprehensive and challenging German benchmark from two major social media platforms that reflects a range of real-world use cases and enables extensive evaluation under language-specific conditions, thereby addressing the lack of diverse non-English benchmarks.

%% file: text/3_data_curation.tex
\section{Data Curation}
\label{sec:data_curation}
For this study, we introduce two large German-language source corpora derived from Reddit and Twitter. From these corpora, we construct five benchmark datasets for AV, covering in-domain, cross-domain, profile-based, and mixed-source settings (see Appendix \ref{sec:gerav_examples} for examples of each dataset and their English translations). The datasets are derived from Twitter and Reddit posts and therefore consist of relatively short texts compared to existing benchmark sources, e.g., derived from novels \cite{bogdanova2014cross}, Wikipedia articles \cite{israeli2025million}, or news \cite{liu2011reuter5050}. 
This characteristic makes the datasets particularly suitable for training systems intended for forensic applications, in which AV is applied to online forums (e.g., incriminated darknet forums or Telegram groups) to identify an active author within a given forum. In the following, the curation and preprocessing steps of each dataset are described.

\paragraph{Reddit \shortname}
We collect a new German Reddit corpus aiming at covering German-language subreddits as comprehensively as possible. Starting from a list of German seed subreddits, we retrieved all posts and iteratively expanded the dataset by following linked usernames to other subreddits, assuming that users active in German communities are likely to participate in additional German-language subreddits.

Using the \textit{langid} \footnote{https://pypi.org/project/langid/} Python package, we retained only posts classified as German, yielding 787,372 posts from 120,538 users across 182 subreddits, being nearly ten times bigger than the German part of the Reddit dataset by \citet{murauer2019generating}. Bot accounts are filtered by excluding usernames containing certain keywords and users with unusually high post counts. To enable pair construction, users with fewer than two posts are removed. To focus on meaningful stylistic features, only posts with a minimum of 25 words are retained. Posts exceeding 1,000 words, containing URLs or user mentions are excluded to reduce noise.

From this corpus, we derive three benchmark AV datasets: in-domain, cross-domain, and profile-based. For all datasets, authors are partitioned into training (60\%), validation (20\%), and test (20\%) splits with no author overlap. For each author, up to two positive pairs (two posts by the same author) and two negative pairs (a post paired with one from a different author in the same split) are sampled. Pair limits keep dataset sizes manageable for training large models, reducing computational cost and training time. We enforce class balance by ensuring an equal number of positive and negative pairs in each split, resulting in balanced binary classification datasets.

To construct the \textbf{reddit cross-domain dataset (rcd)}, we leverage topic separation across subreddits by prompting GPT4o \cite{openai2024gpt4ocard}, resulting in 14 distinct topical domains that were checked manually for plausibility (see Appendix \ref{sec:domain_clustering} for the prompt and \ref{sec:topcial_domains} for the resulting clusters).  In the \textbf{reddit in-domain dataset (rid)}, paired posts always come from the same domain, while in the cross-domain dataset, pairs are drawn from different domains. This cross-domain benchmark creates a more challenging evaluation, which both increases AV difficulty \cite{kestemont2020overview} and reflects real-world applications. Furthermore, this setting enables the investigation of content bias, assessing whether models rely on topical similarity or genuinely capture stylistic signals. Strong performance in topically unrelated settings provides evidence that the models are learning stylistic characteristics rather than merely exploiting content similarities.

Finally, the \textbf{reddit profile-based dataset (rpb)} is created by concatenating all posts from each user in random order into a post list. For each user, a random split point is selected at the end of a post, falling between 30\% and 70\% of the total words in their post list. 
Negative pairs are sampled to ensure that the length difference distributions of positive and negative pairs is similar, thereby minimizing potential length-based biases that a model could exploit. This dataset enables evaluation at the user-profile level, reflecting a realistic scenario in which all available texts of an author are leveraged to make more reliable predictions.

\paragraph{Twitter \shortname} 
The \textbf{Twitter dataset (tw)} is created from \citet{nane_kratzke_2023_7528718}, which contains monthly German Twitter updates from April 2019 to December 2022, and is accessible to researchers upon request. To further ensure the selection of German-language content, \textit{langid} is applied to all texts, resulting in a dataset of 606,588 posts from 54,544 unique authors. In line with the Reddit dataset, users with fewer than two posts are removed, and tweets containing fewer than 25 words are excluded. To mitigate the effect of outliers, tweets exceeding 300 characters and users with more than 200 tweets in total are excluded. 

Consistent with the Reddit \shortname\ dataset, the final Twitter GerAV dataset is constructed by creating training, validation, and test splits with no author overlap. For each author, up to two positive and two negative samples are generated, resulting in a balanced dataset.

\paragraph{Mixed \shortname}
As a final dataset, we combine the three Reddit datasets (rid, rcd, and rpb) and the Twitter dataset by sampling 20,000 pairs for the training set and 4,000 each for validation and test sets from each source while maintaining a balanced distribution of labels, resulting in a \textbf{mixed dataset (mix)}. For an evaluation of GPT-5, we further create a small version of the mixed dataset that covers 480 samples (120 per dataset). 

Table \ref{table:data_stat_comp} summarizes dataset statistics, including the number of samples, unique posts and users, and mean sample length for each dataset. The post length distributions for the Reddit and Twitter datasets before preprocessing are provided in Appendix \ref{sec:gerav_post_lengths}. Both datasets are dominated by short posts, with frequencies decreasing rapidly as length increases. Twitter posts are capped at roughly 75 words, reflecting its maximum post length, while Reddit exhibits a long-tailed distribution.

\begin{table}[t]
\centering
\scalebox{0.9}{
\begin{adjustbox}{width=1\linewidth}
\begin{tabular}{lrrrr} 
\toprule
 \makecell{Dataset} & \makecell{Sample Number} & \makecell{Unique Posts} & \makecell{Unique Users} & \makecell{Mean Sample \\ Len (in Words)} \\ 
\midrule 
Reddit in-domain & 62,328 & 87,979 & 23,083 & 69 \\
Reddit cross-domain & 62,328 & 82,902 & 22,577 & 67 \\ 
Reddit profile-based & 46,166 & 50,775 & 23,083 & 326 \\ 
Twitter & 120,858 & 133,329 & 35,181 & 34 \\ 
Mixed & 112,000 & 158,561 & 47,816 & 124\\
Mixed (small) & 480 & 952 & 692 & 112 \\
\bottomrule
\end{tabular}
\end{adjustbox}
}
\caption[Statistical Comparison: All datasets]{Statistical comparison of all GerAV datasets.}
\label{table:data_stat_comp}
\end{table}

%% file: text/4_Experiment_Setup.tex
\section{Experiment Setup}
\label{sec:experiment_setup}
In this section, we describe the evaluation measures we use and the methods that we benchmark for German AV. This includes (1) established and recent baselines, (2) zero-shot LLMs, and (3) LLMs with LoRAs \cite{hu2022lora} fine-tuned on our training sets. 

\paragraph{Evaluation Measures}
To evaluate the AV methods on \shortname, we measure accuracy in correctly identifying which messages are written by the same or different authors. Further, we measure the F1-score to identify the methods with the highest precision and recall. As a third measure commonly used for AV evaluation, we choose ROC AUC\footnote{\url{https://scikit-learn.org/stable/modules/generated/sklearn.metrics.roc_auc_score.html}} that quantifies how well similarity/distance scores returned by the metric can separate samples from same and different authors with arbitrary thresholds. Note that the first two measures use binary labels, while ROC AUC requires numeric scores. To measure statistical significance, we perform paired non-parametric bootstrap tests for model comparison \citep{efron_bootstrap_1979}. We generate 10,000 bootstrap resamples per dataset and compute F1 scores for each model on identical resampled labels, yielding paired F1 distributions. For each model pair, the p-value was estimated as the proportion of samples in which the higher-scoring model performed no better than its competitor, providing a direct measure of whether observed differences can be attributed to sampling variability.

\paragraph{AV Baselines}
To cover a diverse set of approaches, we start by following \citet{tyo2023valla}, who identify four high-performing AV methods with proven effectiveness \cite{fabien2020bertaa, murauer2021developing, kestemont2020overview, neal2017surveying}. The first method \textit{ngram} is a
feature-based approach by \citet{weerasinghe2021feature}, which represents texts as n-gram character feature vectors and feeds them into a logistic regression classifier. The second baseline, \textit{Prediction by Partial Matching (ppm)} \citet{teahan2003using}, builds a character-level language model for one text and computes cross-entropy on the second text using a ppm compression model. The third method, \textit{hlstm}, introduced by \citet{boenninghoff2019explainable}, employs a hierarchical Bi-LSTM to process text at the character, word, and sentence levels, producing unified style representations for AV. The final baseline \textit{sbert} of this set uses BERT as the backbone for a siamese network trained with contrastive loss. We adapt the implementations to suit German inputs, e.g, BERT has been substituted with its German version \cite{german-bert}. We train each baseline on all five newly created \shortname\ datasets, resulting in a total of 20 models.

Furthermore, we evaluate the recent style embedding models \textit{mStyleDistance} \cite{qiu-etal-2025-mstyledistance} that includes German in its training data and \textit{Multilingual Style Representation} (we abbreviate it as \textit{msr}) \cite{kim-etal-2025-leveraging}, which is multilingual but was not trained on German data. For AV, the cosine similarity between styles is considered. 

To evaluate the baseline models using accuracy and F1 score, we first determine a decision threshold $t$.\footnote{Scores above the threshold indicate ``same author'' while scores below the threshold indicate ``different author''.} The threshold is tuned on the respective validation sets using Youden’s J statistic \cite{youden1950index}: we compute the ROC curve over the validation data and choose the threshold that maximizes the sum of sensitivity and specificity.

\paragraph{Zero-Shot LLMs}
We also benchmark the zero-shot performance of four recent LLMs: \textit{gemma\_3\_12b\_it} \citep{gemma_2025}, \textit{llama-3.1-8b-instruct} \citep{meta2024llama3.1}, \textit{llama-3.2-3b-instruct} \citep{meta2024llama3.2}, and \textit{qwen-2.5-7b-instruct} \citep{qwen2025qwen25technicalreport} using the following prompt template:
\begin{quote}
\textit{Are the following two texts written by the same author?}\newline
\textit{Text A:} \{text\_a\}\newline
\textit{Text B:} \{text\_b\}\newline
\textit{Please answer with ``Yes'' or ``No''.}\newline
\textit{Answer:}
\end{quote}

During inference, we generate tokens until the first ``yes''-token (Yes, yes, YES) or ``no''-token (No, no, NO) is generated. Then, to get a more accurate confidence estimate, we calculate the prediction score as $s = p(\mathit{``yes''-token}) - p(\mathit{``no''-token})$ where we consider the 10 most likely tokens. 
We use $s$ as the input score for the ROC AUC measure. 
Positive values of $s$ mean that the probability of two texts being written by the same author is higher than by different authors. Again, we tune the threshold on the validation set.

As an additional zero-shot model, we evaluate and compare the closed-source model GPT-5 \citep{openai2025gpt5} on 120 subsamples of each dataset besides mixed. For GPT-5, we (1) use the same zero-shot prompt as above and (2) also evaluate a prompt using the LIP prompt component \cite{huang2024can} that instructs the model to first analyze stylistic features of both texts, as well as (3) a version of LIP translated to German (see Appendix \ref{sec:AV prompts} for prompt details).

\paragraph{Tuned LLMs}
We also train LoRAs on each of the four LLMs (\textit{gemma\_3\_12b\_it}, \textit{llama-3.1-8b-instruct}, \textit{llama-3.2-3b-instruct}, and \textit{qwen-2.5-7b-instruct}). For each of our five datasets (Twitter, Reddit in-domain, Reddit cross-domain, Reddit profile-based and Mixed), we train one LoRA per LLM, yielding 20 LoRAs in total. The input prompt follows the zero-shot format but omits ``Please answer with ``Yes'' or ``No.'', as the required output format will be learned during tuning. Information about the training hardware, hyperparameters, runtime and licences can be found in Appendix \ref{sec:training_parameters}. Because the models are tuned on the training sets, we do not perform threshold tuning and use 0 as a threshold for accuracy and F1-score.

%% file: text/5_results.tex
\section{Experiment Results}
\label{sec:experiment_results}

Figure \ref{fig:alias_results_f1} shows the F1-scores of the baselines and tuned LLMs on our datasets (see Appendix \ref{sec:acc_report} for the full results). We first consider the overall performance of the models. We then evaluate the effect of different data sources by comparing results on Twitter and Reddit data. Subsequently, we compare results on Reddit in-domain data with cross-domain data. We also evaluate the effect of text length by comparing message-based with profile-based evaluation. Finally, we present the comparison with closed-source models on a separate subset and analyze the effects of reduced training data and cross-lingual evaluation settings.

\begin{figure}[H]
    \centering
    \includegraphics[width=\linewidth]{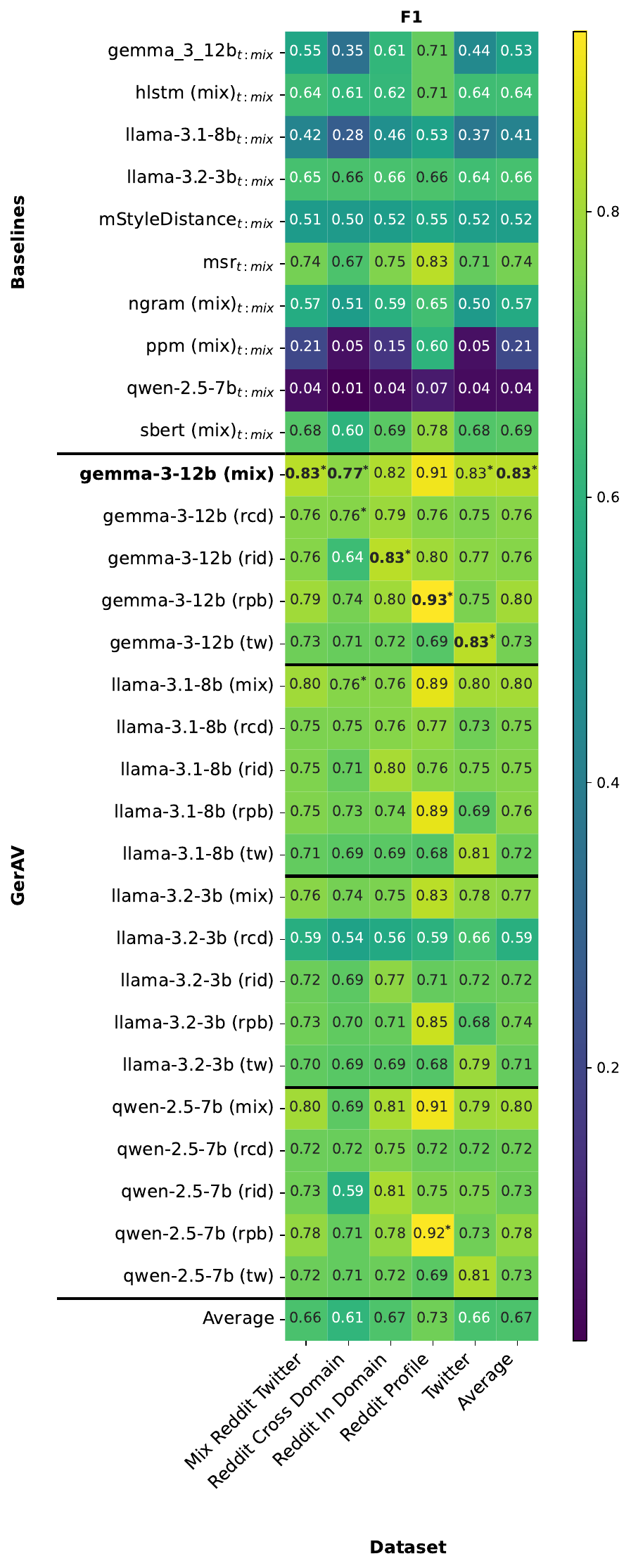}
    \caption{F1-Scores for the baselines and our GerAV models. The y-axis shows the model names and the x-axis shows the evaluated test set. Brackets indicate the training set used: \textbf{Mix} Reddit Twitter (mix), \textbf{R}eddit \textbf{C}ross \textbf{D}omain (rcd), \textbf{R}eddit \textbf{I}n \textbf{D}omain (rid), \textbf{R}eddit \textbf{P}rofile \textbf{B}ased (rpb) and \textbf{Tw}itter (tw). The subscript $t$ denotes the validation set used to tune decision thresholds for the baselines. Column-wise best results are shown in bold. The best-performing model groups for each dataset (defined as those that statistically outperform all other models (p < 0.01) and show no statistically significant differences within the group) are marked with an asterisk next to the score.}
    \label{fig:alias_results_f1}
\end{figure}

\begin{figure}[!ht]
    \centering
    \includegraphics[width=\linewidth]{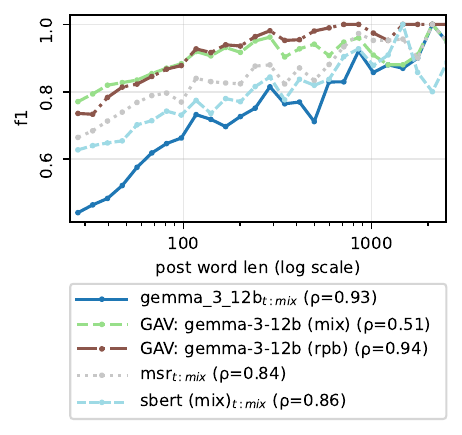}
    \caption{F1-Score per message length in words (up to 2500) on the mixed test set. $\rho$ is the Spearman correlation between message length and F1 scores. We display the results with bucket sizes increasing on a log-scale, to account for less examples existing for longer messages.}
    \label{fig:f1perlen}
\end{figure}

\paragraph{Overall Performance}
We first consider the average (6th) column, finding that models trained on the mixed train set achieve the highest performance. Specifically, Gemma-3 (mix) scores the highest overall F1 of 0.83, 0.03 points above Llama-3.1 (mix) and Qwen-2.5 (mix) (p<0.01). The second-best results come from models trained on the profile-based dataset, with Gemma-3 (rpb) scoring 0.80 and Qwen-2.5 (rpb) 0.02 points lower. For Gemma-3 and Llama-3.1, the lowest average performance occurs when trained on the Twitter dataset, while for Llama-3.2 and Qwen-2.3, it occurs with Reddit cross-domain.

Among the baselines, Multilingual Style Representation (msr), with its threshold tuned on the mixed validation set, achieves the highest average F1 of 0.74, outperforming the second-best baseline, sbert, by 0.05. This also exceeds the best zero-shot baseline, Llama-3.2, which attains an average F1 of 0.66, while hlstm performs nearly on par, trailing by 0.02. Compared to our tuned LLMs, msr matches the average F1 of Llama-3.2 (rpb). The weakest models are Qwen-2.5 in zero-shot, with F1 = 0.04, followed by ppm (mix) with F1 = 0.21. Since the mixed test set contains a stratified subset of the other test sets, its results closely mirror the average column, differing by at most 0.02.

These results show that our fine-tuned LLMs clearly outperform existing AV baselines, and their performance predictably correlates with model size, with larger models generally performing better. This suggests that AV benefits from greater model capacity and that even stronger results may be achievable with larger-scale models. However, parameter efficiency remains important, making msr as the best-performing baseline an attractive alternative when computational resources are limited.

Overall, the strong performance of models trained on the mixed dataset suggests that exposure to stylistic variability across domains enables learning robust and generalizable representations of authorial style, challenging the assumption that models must be highly data-specific. This is particularly relevant in real-world applications where test data may differ from any single training source. 

\paragraph{Reddit vs.\ Twitter}
Here, we consider the evaluation on the Reddit in-domain dataset (3rd column) and Twitter dataset (5th column). Models tuned on the Twitter training set perform better on the Twitter test set (Gemma: +0.11 F1), while models trained on Reddit in-domain perform better on the Reddit in-domain test set (Gemma: +0.06 F1), indicating that models capture dataset-specific characteristics such as message style, topic, or length. The larger gap for Reddit suggests that Reddit data promotes greater generalization than Twitter. The strongest model on Reddit in-domain is Gemma-3 (rid), and the strongest model on Twitter is Gemma-3 (tw), with Gemma-3 (mix) consistently ranking second (non-significant difference on Twitter, p>=0.01), again highlighting the benefit of combining training data from multiple sources. The average performance across all models for Reddit in-domain is 0.01 points higher than for Twitter.

\paragraph{In-Domain vs.\ Cross-Domain}
Next, we compare the performance on the Reddit in-domain test set (3rd column) with the cross-domain test set (4th column). In most cases, models trained on the respective training set (in-domain and cross-domain) perform stronger on the respective test set than vice versa. 
However, for Llama-3.2, training on cross-domain leads to a much lower performance than with in-domain for all dataset  (0.13 lower F1-score on average). Perhaps, because of its smaller parameter count, the model is not able to handle the more complex case of cross-domain evaluation well. Also, the mixed data models outperform or perform on par with the cross-domain models on the cross-domain dataset. This means that mixing data from many sources has a higher benefit for cross-domain evaluation than just training on cross-domain data acquired in the same manner as the test set. The average across all models shows that cross-domain evaluation is more difficult than in-domain evaluation, with the average F1-score being 0.06 higher for in-domain. Overall, the slight performance drop suggests that the models do rely to some extent on content cues. However, the modest decrease indicates that they still capture meaningful stylistic signals and remain effective under content-controlled conditions.

\paragraph{Message-Based vs.\ Profile-Based}
Here, we compare the performance on the Reddit in-domain test set (3rd column) with the profile-based test set (5th column). As before, models trained on the respective datasets show the strongest performance, with Gemma-3 (rpb) and Qwen-2.5 (rpb) performing best. The mixed models perform almost on par for both datasets. For example, for profile-based evaluation, Gemma-3 (mix) has only 0.02 difference in F1-Score. Overall, profile-based evaluation yields higher metrics than other scenarios (e.g., 0.09 higher accuracy), likely because concatenated messages provide more information on an author’s style than single messages. The best-performing model remains a tuned Gemma variant.

Figure \ref{fig:f1perlen} shows F1-scores for the two best-performing baselines, msr and sbert (trained on mixed), as well as Gemma-3-12b in zero-shot and fine-tuned (mixed and profile-based) settings across varying word counts (x-axis), alongside corresponding Spearman correlations. To this end, samples from the mixed test set were grouped into word-length buckets on a logarithmic scale (see Appendix \ref{sec:bins} for details). Gemma-3 (mix) performs best up to about 100 words, after which it is surpassed by Gemma-3 (rpb). Around 500 words, Gemma-3 (mix) reaches a performance plateau, and at about 1000 words, the performance of other models is almost on par with it. Gemma-3 (mix) is decreasing from about 500 words. Gemma-3 zero-shot shows the steepest gain, starting with the lowest F1 and gradually catching up as word count increases. The baseline models exhibit similar curves, with msr generally outperforming sbert except at around 1500 words, where sbert peaks before declining. The high Spearman correlations between message length and F1-Score (up to $\rho$=0.94) indicate that AV is highly sensitive to text length, with short texts often lacking sufficient stylistic information. This highlights the importance of minimum length requirements in practical use.

\paragraph{Closed-Source Evaluation}
\begin{figure}[!ht]
    \centering
    \includegraphics[width=\linewidth]{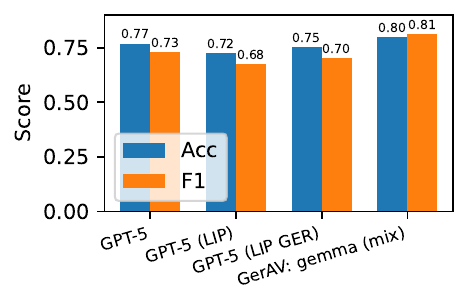}
    \caption{Accuracy and F1-scores of GPT-5 and GerAV: gemma-3-12 (mix). They are evaluated on a stratified 480 sample subset of the mixed test set.}
    \label{fig:gpt_comp}
\end{figure}

Figure \ref{fig:gpt_comp} shows the accuracy and F1-scores of Gemma-3 (mix) on a stratified subset of the mixed test set. The GerAV model significantly outperforms GPT-5 by 0.031 accuracy and 0.081 F1-score ( p<0.01). ROC scores are not reported, as GPT-5’s log probabilities are inaccessible. This result is especially significant for applications, as fine-tuned open-source LLMs can be deployed locally, providing full control over data and reproducibility without sacrificing performance. Interestingly, GPT-5 with the short zero-shot prompt achieves a better performance than with both LIP prompts, with the German LIP prompt performing slightly better than the English one.

\paragraph{Effect of Number of Training Labels}

\begin{figure}[!ht]
    \centering
    \includegraphics[width=1\linewidth]{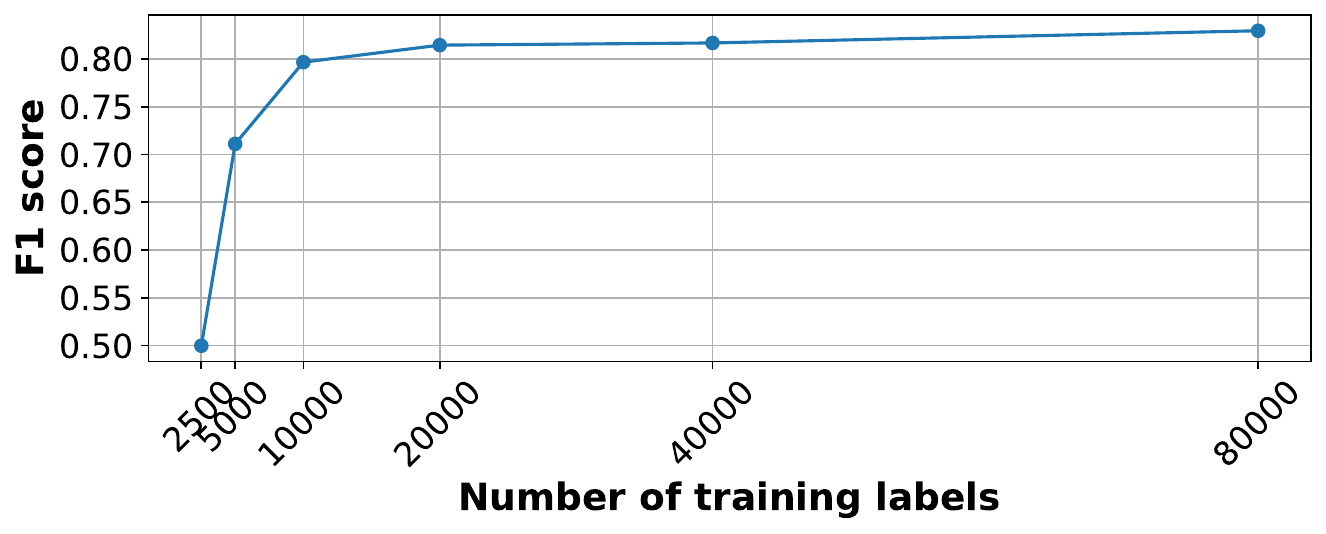}
    \caption{F1 scores of Gemma-3-12b plotted against number of training samples.}
    \label{fig:no_training_samples_vs_f1}
\end{figure}

While GerAV comprises a large number of authors and texts, in realistic scenarios, training data is often scarce, particularly for low-resource languages. To investigate the impact of limited training data, we analyze the effect of progressively reducing the number of training samples during fine-tuning of our best-performing Gemma model. Specifically, we iteratively halved the size of the mixed training set five times and retrained the model on each subset.

Full results are provided in Appendix \ref{sec:varying_training_number}. Figure \ref{fig:no_training_samples_vs_f1} presents the F1 scores when tested on the mixed test set. Reducing the dataset from 80,000 to 70,000 samples has a negligible impact (0.83 to 0.82, p < 0.01). Performance declines gradually to 0.79 with one-eighth of the data, but drops more sharply at smaller scales (0.71 at 5,000; 0.52 at 2,500), indicating that this range approaches a lower limit for effective adaptation. Overall, these findings indicate that training data can be substantially reduced while maintaining competitive performance, with performance gains largely saturating at around one-fourth of the full dataset. This highlights the potential of the proposed LoRA approach for application in low-resource settings.

\paragraph{Language-Specific Investigation}

\begin{table}[t]
\centering
\scalebox{0.9}{
\begin{adjustbox}{width=0.5\textwidth}
\begin{tabular}{lrr} 
\toprule
 \makecell{model} & \makecell{Twitter German} & \makecell{Twitter English}  \\ 
\midrule 
gemma-3-12b (tw\_en) & 0.70 & 0.73 \\
gemma-3-12b (tw\_ger) & 0.82 & 0.68 \\ 
\bottomrule
\end{tabular}
\end{adjustbox}
}
\caption[F1 Scores: Cross-lingual evaluation]{F1-Scores for models trained on English (tw\_en) and German (tw\_ger) twitter datasets and tested on respective test datasets.}\label{table:twitter_english}
\end{table}

To further examine the value of language-specific AV corpora and models, we use an English Twitter dataset from \citet{englishTwitter}, following the splits defined by \citet{tyo2023valla}. For comparability, we applied our preprocessing pipeline, resulting in 22,352 training samples. We then sampled an equally sized subset from GerAV Twitter.

We trained Gemma-3-12b on both datasets and evaluated each model on both test sets (see Table \ref{table:twitter_english}). While the German model reaches an F1 of 0.82 when applied to the German test set, the English model only reaches 0.70 F1 when tested on the English test set, suggesting that the English dataset is more challenging, potentially due to shorter messages (mean length: 26.9 in English vs. 34.1 words in German). Cross-lingual evaluation reveals performance drops: the German model reveals a moderate drop, scoring 0.68 on English, while the English model drops more substantially to an F1 of 0.70 on German test data. Overall, these results indicate that less dominant languages benefit more from language-specific fine-tuning, underscoring the importance of dedicated AV datasets. The dominance of English pretraining appears to partially compensate for task adaptation on a different language, though a notable performance gap remains.

We further evaluated the German LLM LLäMmlein-7b-chat \citep{pfister-etal-2025-llammlein}, to assess whether language-specific pretraining and embeddings provide measurable advantages for AV. We train and test the model on all GerAV subsets but did not observe any performance gains over the English LLMs, achieving e.g. an F1 of 0.78 when trained and tested on the mixed dataset(see Appendix \ref{sec:acc_report} for full results). This suggests that our fine-tuning approach can sufficiently adapt models to language-specific phenomena relevant to AV even when the pretraining data is dominated by another language. However, this finding should be confirmed by evaluating more models and languages in future work.

%% file: text/7_conclusion.tex
\section{Conclusion}
\label{sec:conclusion}
Overall, this work provides a comprehensive evaluation of fine-tuned LLMs alongside a diverse set of established baselines representing different methodological approaches on our newly introduced \shortname\ benchmark. In doing so, we address two major gaps in the AV literature: the lack of systematic evaluations including multiple LLMs and multiple baselines and the scarcity of robust benchmarks for non-English languages. Our results provide several insights into the behavior of LLMs for AV in German with implications for both real-world applications and future research. (1) We clearly demonstrate the superiority of fine-tuned LLMs over traditional baselines. (2) We show that the best-performing model, Gemma-3-12b, is robust to domain shifts and handles both very short and very long input texts well. Moreover, strong performance can be maintained even when the amount of training data is substantially reduced. (3) Our results show the importance of training on language-specific corpora, revealing significant performance drops in cross-lingual settings. (4) \shortname\ demonstrates that training on combined datasets enhances generalization abilities, alleviating the need for highly data-specific models, and provides a solid foundation for future work on improving performance in cross-domain scenarios as well as handling varying text lengths.

Overall, our results provide a solid foundation for future research aimed at improving performance and paving the way toward robust, high-performing AV models suitable for practical applications across languages.

%% file: text/8_limitations.tex
\section*{Limitations}
\label{sec:limitations}
Key limitations of this work include the need for a critical assessment of potential biases in the datasets and the absence of interpretable features in the model outputs. Future work should focus on incorporating explanations into the framework to address the crucial requirement of interpretability in AV systems. In addition, the influence of content has been only partially addressed and should be examined in more detail to ensure that model predictions are driven by authorial style rather than topical cues. The \shortname\ cross-domain dataset should further be extended to more challenging scenarios, such as cross-platform author identification or varying text genres produced by the same author. 

In addition, the evaluation of our fine-tuning strategy is limited, and future work should explore a wider range of architectures, prompt designs, hyperparameter settings, and adaptation methods. The experimental setup should be extended to additional non-English languages beyond German with a specific focus on low-resource languages.

Finally, this work does not include human evaluation or direct comparisons between model and human performance, which are necessary steps toward reliably replacing expert judgment with automated AV systems. The absence of human evaluation in AV tasks is a general issue in the literature and should be addressed in future work.

%% file: text/9_ethics.tex
\section*{Ethical Considerations}
\label{sec:ethical_considerations}

While authorship analysis generally aims to support socially beneficial applications, such as forensic investigations or plagiarism detection, these methods may also be misused: awareness of stylistic markers could enable deliberate style obfuscation, and the analysis of anonymous texts may infringe on privacy or undermine legitimate anonymity. We therefore emphasize that authorship verification results are probabilistic and should be applied cautiously, particularly in sensitive or high-stakes contexts.

We report all licenses and configurations of models used in this work (see Appendix \ref{sec:training_parameters}), and all data used in this work is publicly available for research purposes. Information about data licences and access can be found in Appendix \ref{sec:data_licence}. We did not extract nor try to predict any personally identifiable information, such as e.g, names, birth dates, addresses, or gender, of any author included in the datasets used for this work. While we assume that our developed AV systems do not systematically predict two texts to originate from the same author for reasons unrelated to stylistic similarity, such as shared social or demographic characteristics, we do not investigate this behavior further. Consequently, we cannot guarantee the fairness of our models.  

%% file: text/10_acknowledgements.tex
\section*{Acknowledgements}
\label{sec:acknowledgements}
We gratefully acknowledge support from the German Federal Ministry of Research, Technology and Space (BMFTR) through the research grant ALiAS (13N17272 and 13N17273) in the security research program, and from the German Research Foundation (DFG) through the Heisenberg Grant EG 375/5-1.

%% file: text/11_appendix.tex
\section{GerAV Examples}
\label{sec:gerav_examples}
Table \ref{table:data_type_examples} shows examples for each dataset in GerAV and their English translation.

\begin{table*}[]
\small
\centering
\begin{tabular}{p{0.06\linewidth}p{0.425\linewidth}p{0.425\linewidth}}
\toprule
\textbf{Dataset} & \textbf{Text A} & \textbf{Text B} \\
\midrule
\multicolumn{2}{l}{\textbf{Original examples}}\\ \cmidrule{1-2}
Twitter & Das sieht man z.B. an
München, hier verkauft die Stadt eine Umweltspur als Paradigmenwechsel, doch in Wirklichkeit müssen hier Radler die Spur mit Bussen und E-SUV teilen - kein Fortschritt & Ist hier irgendwo ein*e Arzt*in aus \#
Stuttgart? Das ist ein tolles Projekt, das sich zu unterstützen lohnt. Es gibt Malteser Medizin übrigens auch in \#
Nürnberg\\
\midrule
Reddit In-Domain & \textbf{Subreddit: wasistdas} | Sieht nach Weihnachtsbeleuchtung aus  Da gibt es recht viele verschiedene Lämpchen   Die "Watt Zahl " hängt unter anderem davon ab wie viele Lampen an der Lichterkette sind &\textbf{Subreddit: wasistdas} | Das müsste zum Testen von Batterien sein  Zumindest erinnere ich mich an so einen Widerstands- Draht in dem Zusammenhang   Und 2 volt passt für die einzelnen Zellen einer Batterie vom Gabelstapler usw \\
\midrule
Reddit Cross-Domain & \textbf{Subreddit: Ratschlag} | Es gibt aus dem Material und damit aus genau der Stärke vom Stoff auch Mützen. Buff hat welche, die sind extra fürs Tragen unterm Helm gemacht, also ultra dünn und trotzdem halten sie den Wind fern. & \textbf{Subreddit: kannmandasnochessen} | Hab in dem Monat so eine Dose mit 2013 drauf im Keller gefunden. Aufmachen, riechen, kosten. Meine war genauso wie sie auch vor 13 Jahren geschmeckt hätte. \\
\midrule
Reddit Profile-Based & [...]Finde ich top, aber da gibts auch viele andere gute Alternativen.   Sonst finde ich irgendeine Art v[...] & [...]en, nach meinem (rein subjektiven) Gefühl ist das von Landkreis zu Landkreis anders. Grundsätzlich h[...] \\
\midrule
\multicolumn{2}{l}{\textbf{Translated examples}}\\ \cmidrule{1-2}
Twitter & 
This can be seen in Munich, for example, where the city is promoting a bike lane as a paradigm shift, but in reality cyclists have to share the lane with buses and electric SUVs—no progress at all. & Is there a doctor from \#Stuttgart here? This is a great project that is worth supporting. Incidentally, Malteser medical services are also available in \#Nuremberg.\\
\midrule
Reddit In-Domain & \textbf{Subreddit: whatisthat} | Looks like Christmas lights. There are quite a few different lights. The “wattage” depends, among other things, on how many lights are on the string. &\textbf{Subreddit: whatisthat} | That must be for testing batteries. At least, I remember seeing a resistance wire like that in that context. And 2 volts is suitable for the individual cells of a forklift battery, etc. \\
\midrule
Reddit Cross-Domain & \textbf{Subreddit: Advice} | There are also hats made from this material, which means they have the same thickness as the fabric. Buff has some that are specially designed to be worn under a helmet, so they are ultra-thin but still keep the wind out. & \textbf{Subreddit: canyoustilleatthat} | I found a can with 2013 on it in the basement this month. Open it, smell it, taste it. Mine tasted exactly as it would have 13 years ago. \\
\midrule
Reddit Profile-Based & [...]I think it's great, but there are also many other good alternatives.   Otherwise, I find some kind o[...] & [...]en, in my (purely subjective) opinion, this varies from county to county. Basically, h[...] \\
\bottomrule
\end{tabular}
\caption{Examples of our four sub-datasets with same-author pairs and their English translations. The mixed dataset is excluded, as is built from these four. For profile-based, we only show a small part of the concatenated messages. We changed the involved entities.}
\label{table:data_type_examples}
\end{table*}

\section{Domain Clustering Prompt}
\label{sec:domain_clustering}
Figure \ref{fig:cluster_prompt} shows the prompt used to prompt GPT-4o for clustering German subreddits into topical domains. 

\begin{figure}[h!]
\begin{tcolorbox}[width=\linewidth, arc=10pt, boxrule=0.3mm]
I will provide you with a list of German subreddits. Your task is to cluster these subreddits into semantically coherent topical groups. Create clusters that are as distinct from each other as possible. You may ignore subreddits that don’t fit any cluster. 

For each cluster, provide: 
\begin{itemize}[itemsep=0.5pt, topsep=2pt]
\ttfamily
    \item A cluster name 
    \item A short description of the theme 
    \item The list of subreddits belonging to it 
\end{itemize}

If a subreddit could fit into multiple topics, place it in the one that fits best and explain your choice briefly. Do not create clusters just to use all subreddits; fewer, cleaner clusters are preferred. 

Here is the subreddit list:

[...]
\end{tcolorbox}
\caption{Prompt used for clustering German Subreddits into topical domains.}
\label{fig:cluster_prompt}
\end{figure}

\section{Topical Domains}
\label{sec:topcial_domains}
Table \ref{table:subreddit_clusters} displays the names of all topical domains, a description of each domain and all subreddits belonging to this domain as outputted by GPT-4o.

\begin{table*}[!ht]
\footnotesize
\begin{adjustbox}{width=\textwidth}
\begin{tabular}{p{3cm} p{3.5cm} p{6.5cm} p{1.5cm}}
\toprule
\makecell{Cluster} & \makecell{Description} & \makecell{Subreddits} & \makecell{Counts} \\
\midrule
General German Community \& Discussion & Broad discussion spaces about life in Germany, general questions, culture, and mixed topics. & de, AskGermany, FragReddit, FragenUndAntworten, KeineDummenFragen, Ratschlag, wirklichgutefrage, tja, WerWieWas, de\_IAmA, Nachrichten, GuteNachrichten, DErwachsen, einfach\_posten, wasistdas, serviervorschlag, duschgedanken, tragedeigh & 110,453\\
\midrule
Humor, Memes \& Satire & German memes, ironic content, absurd humor, and satire. & ich\_iel, hessich\_iel, LTB\_iel, deutschememes, Berliner\_memes, aberBitteLaminiert, asozialesnetzwerk, OkBrudiMongo, schkreckl, wortwitzkasse, wasletztestern, wasletzterezension, NichtDieTagespresse, satire\_de\_en, GermansGoneWild, ichbin14unddasisttief & 47,302 \\
\midrule
Politics, Society \& Ideology & German politics, political parties, activism, and social debates. & politik, ich\_politik, PolitikBRD, DIE\_LINKE, Kommunismus, antiarbeit, Klimawandel, Klimagerechtigkeit, umwelt\_de, Verkehrswende, OeffentlicherDienst, GoldenerAluhut & 44,741\\
\midrule
Finance, Careers \& Economics & Personal finance, investing, macroeconomics, taxes, and professional life. & Finanzen, Aktien, Immobilieninvestments, Normalverdiener, Wirtschaftsweise, Energiewirtschaft, Steuern, arbeitsleben, Azubis, InformatikKarriere, selbststaendig, mauerstrassenwetten, CapitolVersicherungAG & 92,212 \\
\midrule
Technology, IT \& Engineering & Computing, hardware, electronics, IT knowledge, and advanced technologies. & de\_EDV, PCGamingDE, PCBaumeister, Elektroinstallation, technologie, KI\_Welt, informatik & 19,824\\
\midrule
Transportation, Mobility \& Vehicles & Cars, bikes, trains, aviation, and everyday mobility topics. & Fahrrad, automobil, autobloed, TuningVerbrechen, RoestetMeinAuto, Elektroautos, MotorradDeutschland, deutschebahn, bahn, LuftRaum, Fuehrerschein, Falschparker & 46,116\\
\midrule
Food, Cooking \& Household & Everyday cooking, recipes, food culture, organization, and home life. & Kochen, Backen, keinstresskochen, vegetarischDE, VeganDE, kantine, kannmandasnochessen, SchnitzelVerbrechen, Leberkasverbrechen, Doener, Doenerverbrechen, SpeziVerbrechenDE, Kleiderschrank, Einrichtungstipps, wohnkultur, wohnen, Hausbau, Canbau, Garten, Balkonkraftwerk & 84,548 \\
\midrule
Regional \& Local Communities & City-based, state-based, and regional identity communities in German-speaking areas. & frankfurt, Leipzig, dresden, karlsruhe, bavaria, wien, VfBStuttgart, BayernMunich, borussiadortmund, eintracht, MannausSachsen, RentnerfahreninDinge, rentnerzeigenaufdinge & 36,169 \\
\midrule
Science, Knowledge \& Education & Academic learning, research, history, science, documentaries, and reading. & Studium, abitur, WissenIstMacht, Wissenschaft, Weltraum, Geschichte, GeschichtsMaimais, Dokumentationen, buecher & 34,202\\
\midrule
Lifestyle, Relationships \& Personal Life & Relationships, parenting, gender discussions, fitness, self-expression. & beziehungen, FragtMaenner, FragNeFrau, Eltern, Weibsvolk, FitnessDE, BeautyDE, Beichtstuhl, BinIchDasArschloch & 84,948 \\
\midrule
Pets, Animals \& Nature & Pets, wildlife, nature exploration, and animal-related humor. & Katzengruppe, Gittertiere, naturfreunde, PferdeSindKacke & 7,404\\
\midrule
Media, Entertainment \& Pop Culture & Movies, music, podcasts, creators, streaming, and entertainment content. & Filme, musik, YouTubeDE, Twitch\_DE, FestundFlauschig, WolfgangMSchmitt, Augenschmaus, Augenbleiche, zocken, de\_punk & 23,979\\
\midrule
Work, Trades \& Practical Skills & Craftsmanship, DIY skills, emergency services, manual professions. & Handwerker, Handarbeiten, selbermachen, feuerwehr & 10,887\\
\midrule
Law, Administration \& Institutions & Legal issues, police, military, emergency services, and administrative institutions. & recht, polizei, pozilei, blaulicht, bundeswehr, dhl\_deutsche\_post & 20,476 \\
\bottomrule
\end{tabular}
\end{adjustbox}
\caption[German subreddit clusters]{Table showing clusters of German subreddits, including a brief description, subreddit membership, and post counts.}
\label{table:subreddit_clusters}
\end{table*}

\section{GerAV Post Lengths}
\label{sec:gerav_post_lengths}
\begin{figure}[!ht]
    \centering
    \includegraphics[width=1\linewidth]{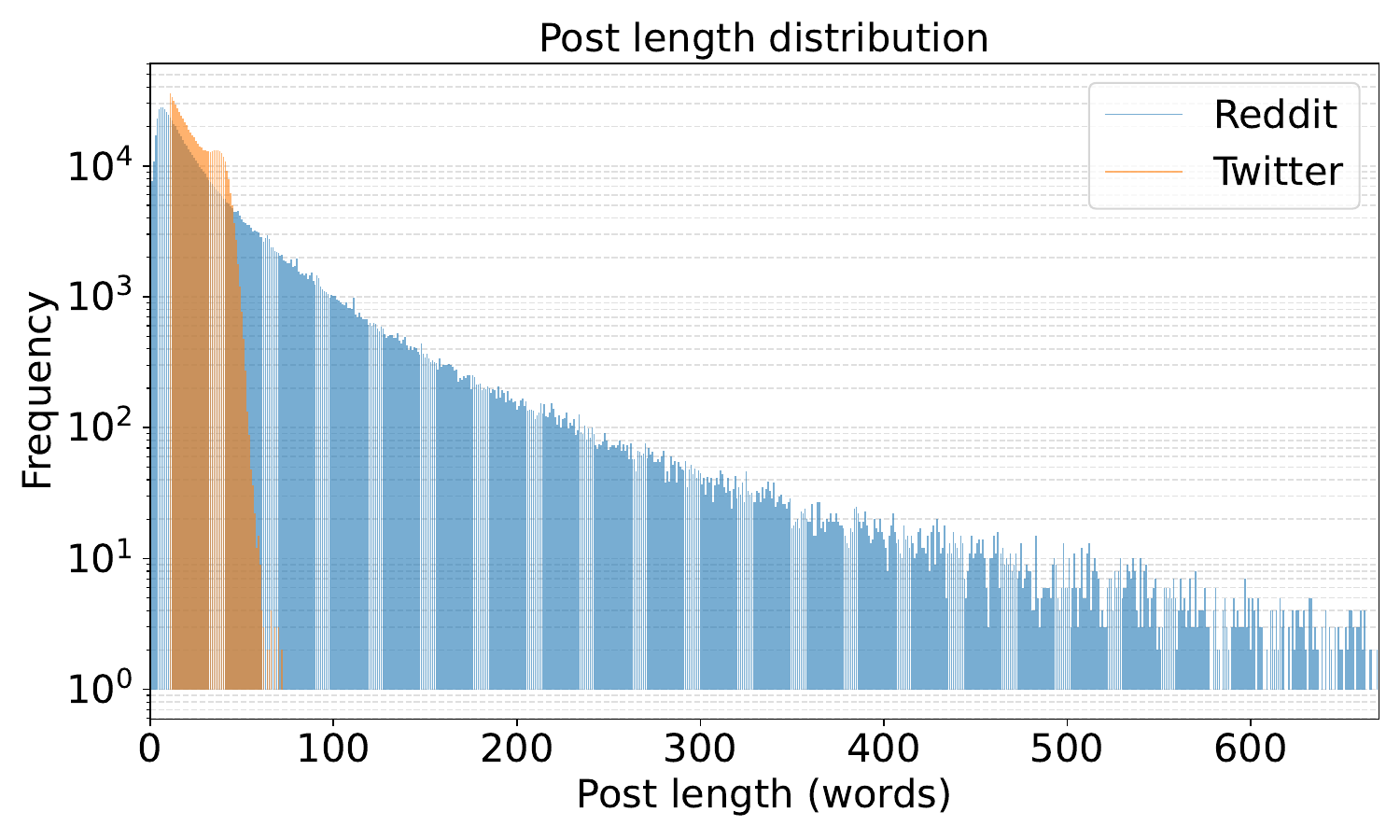}
    \caption{Log frequencies of post word lengths for the Reddit (blue) and Twitter (orange) dataset before preprocessing. The long tail of the Reddit distribution is truncated at 0.05\% for visualization purposes; the true maximum post length is 4,321 words.}
    \label{fig:post_len_comparison}
\end{figure}

Figure \ref{fig:post_len_comparison} shows the distributions in the German Reddit and Twitter datasets before preprocessing. 

\newpage
\section{AV Prompts}
\label{sec:AV prompts}
Below, we list all prompts that were used for the AV experiments in this paper:
\begin{enumerate}
    \item For LLM baselines (including GPT-5):
    \begin{quote}
\textit{Are the following two texts written by the same author?}\newline
\textit{Text A:} \{text\_a\}\newline
\textit{Text B:} \{text\_b\}\newline
\textit{Please answer with ``Yes'' or ``No''.}\newline
\textit{Answer:}
\end{quote}
    \item For fine-tuned LLMs:
    \begin{quote}
\textit{Are the following two texts written by the same author?}\newline
\textit{Text A:} \{text\_a\}\newline
\textit{Text B:} \{text\_b\}\newline
\textit{Answer:}
\end{quote}
    \item For the English LIP prompt, slightly adjusted to prevent the LLM from refusing the task:
    \begin{quote}
\textit{For a scientific experiment, given two texts determine if they are written by the same author. Analyze the writing styles of the input texts, disregarding the differences in topic and content. Focus on linguistic features such as phrasal verbs, modal verbs, punctuation, rare words, affixes, quantities, humor, sarcasm, typographical errors, and misspellings.}\newline
\textit{Text A:} \{text\_a\}\newline
\textit{Text B:} \{text\_b\}\newline
\textit{Please think step-by-step, then answer with 'Yes' or 'No' as your last word.}\newline
\textit{Answer:}
\end{quote}
\item For the German translation of the LIP prompt:
\begin{quote}
\textit{Für ein wissenschaftliches Experiment möchten wir feststellen, ob zwei Texte von demselben Autor verfasst wurden. Analysiere die Schreibstile der Eingabetexte und ignoriere dabei Unterschiede im Thema und Inhalt. Konzentriere dich auf linguistische Merkmale wie Phonologie, Morphologie, Wortbildung, Syntax, Wortstellung, Kasussystem, Genus, Tempus, Modus, Passiv, Kongruenz, Valenz, Artikelgebrauch, Pronomen, Negation, Modalverben, Semantik, Pragmatik, Informationsstruktur, Prosodie, Wortarten, Nebensätze, Konjunktiv, Modalpartikeln, seltene Wörter und Idiomatik.}\newline
\textit{Text A:} \{text\_a\}\newline
\textit{Text B:} \{text\_b\}\newline
\textit{Bitte denke Schritt für Schritt nach und antworte am Ende mit 'Ja' oder 'Nein' als letztes Wort.}\newline
\textit{Answer:}
\end{quote}
\end{enumerate}

\section{Training Setup, Hyperparameters and Model Licences}
\label{sec:training_parameters}
We train the LoRAs on a Slurm cluster with H100 GPUs. Training times range from under an hour for Llama3.2-3B to 4:40h for Gemma-12B on the mixed dataset. We use the following hyperparameters (evaluated in experiments on the Twitter validation set): batch\_size=2, num\_train\_epochs=1, tuning\_seed=10, lang="en", learning\_rates=3e-4, weight\_decays=0.01, seeds=42, do\_lora=true, lora\_rs=128, lora\_alphas=32, lora\_dropouts=0.0. We use the following library versions: \textbf{torch 2.9.0} \cite{Ansel_PyTorch_2_Faster_2024}, 
\textbf{transformers 4.57.3} \cite{Wolf_Transformers_State-of-the-Art_Natural_2020}, 
\textbf{vllm 0.12.0} \cite{kwon2023efficient}, 
\textbf{trl 0.21.0} \cite{von_Werra_TRL_Transformers_Reinforcement_2020}.

 Our use of all pretrained language models is consistent with their intended use as specified in the respective licenses that are listed in Table \ref{table:model_licences}. The models were used solely for research purposes, including evaluation and fine-tuning, and in compliance with all stated access and usage conditions. Any derived artifacts created in this work are explicitly intended for research use only, and their intended use remains compatible with the original license terms and downstream restrictions of the source models.

\begin{table*}[!ht]
\centering
\begin{tabularx}{\textwidth}{p{2.8cm} p{4cm} X}
\toprule
\textbf{Model} & \textbf{License / Terms} & \textbf{Link} \\
\midrule
Gemma-3-12B-it \cite{gemma_2025} & Gemma Terms of Use & \url{https://ai.google.dev/gemma/terms} \\
LLaMA-3.1-8B-instruct \cite{meta2024llama3.1} & LLAMA 3.1 Community License Agreement & \url{https://github.com/meta-llama/llama-models/blob/main/models/llama3_1/LICENSE} \\
LLaMA-3.2-3B-instruct \cite{meta2024llama3.2} & LLAMA 3.2 Community License Agreement & \url{https://huggingface.co/meta-llama/Llama-3.2-3B-Instruct} \\
LLäMmlein-7B-chat \cite{pfister-etal-2025-llammlein} & LLäMmlein RESEARCH-ONLY RAIL-M & \url{https://huggingface.co/LSX-UniWue/LLaMmlein_7B_chat/blob/main/license.md} \\
Qwen-2.5-7B-instruct \cite{qwen2025qwen25technicalreport} & Apache License & \url{https://huggingface.co/Qwen/Qwen2.5-7B-Instruct/blob/main/LICENSE} \\
\bottomrule
\end{tabularx}
\caption{\raggedright License names and links to all models used for fine-tuning in this work.}
\label{table:model_licences}
\end{table*}

\section{Accuracy, F1 and RoC\_AuC on \dataset}
\label{sec:acc_report}
\begin{figure*}[t]
    \centering
    \includegraphics[width=\linewidth]{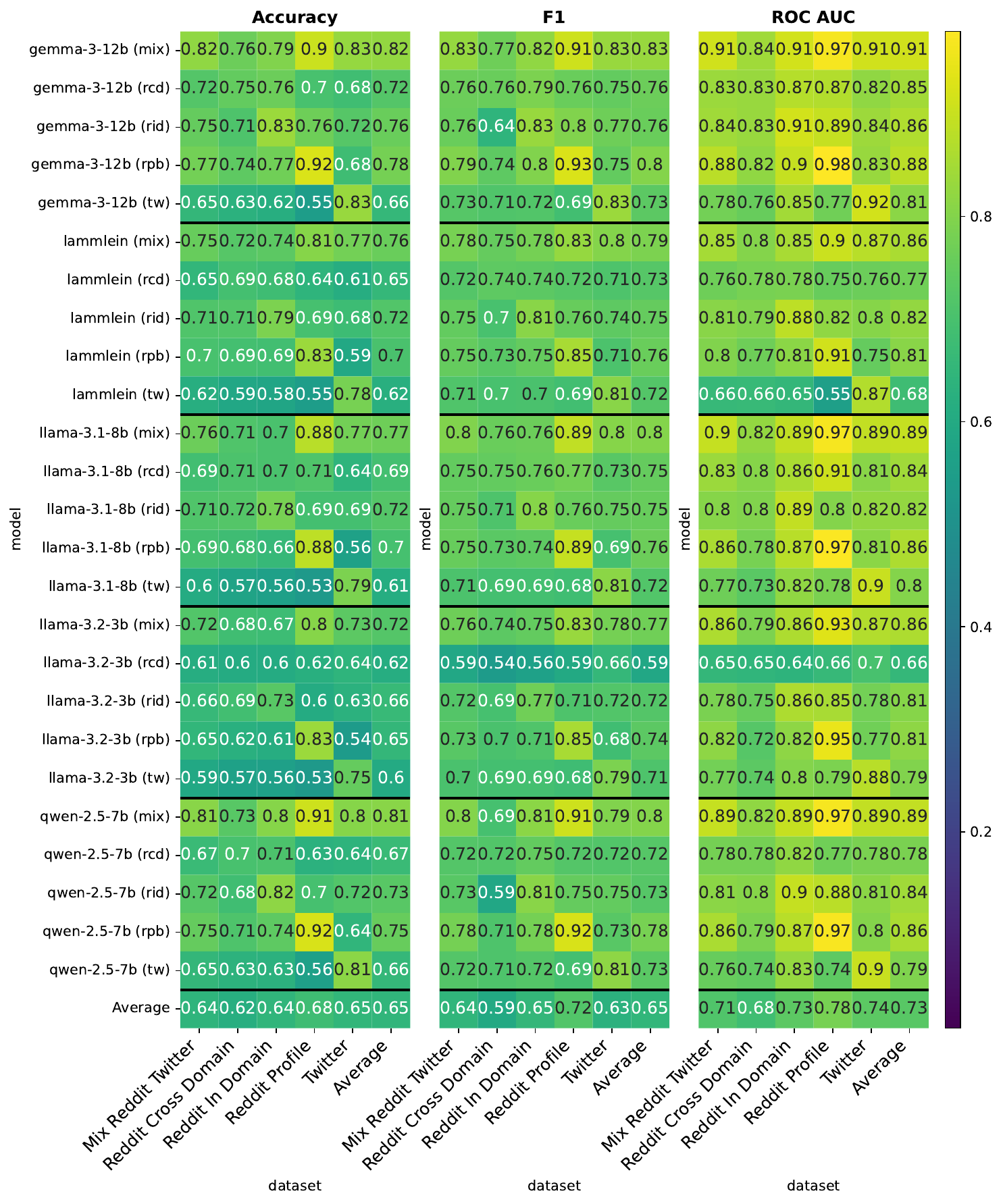}
    \caption{Accuracy, F1-Score and RoC\_AuC Score for the baselines. The y-axis shows the model names and the x-axis shows the evaluated test set. The subscript indicates the validation set that was used to tune the decision threshold: \textbf{Mix} Reddit Twitter (t:mix), \textbf{R}eddit \textbf{C}ross \textbf{D}omain (t:rcd), \textbf{R}eddit \textbf{I}n \textbf{D}omain (t:rid), \textbf{R}eddit \textbf{P}rofile \textbf{B}ased (t:rpb) and \textbf{Tw}itter (t:tw). The highest value in every column is written in bold.}
    \label{fig:alias_results_baslines}
\end{figure*}

\begin{figure*}[t]
    \centering
    \includegraphics[width=\linewidth]{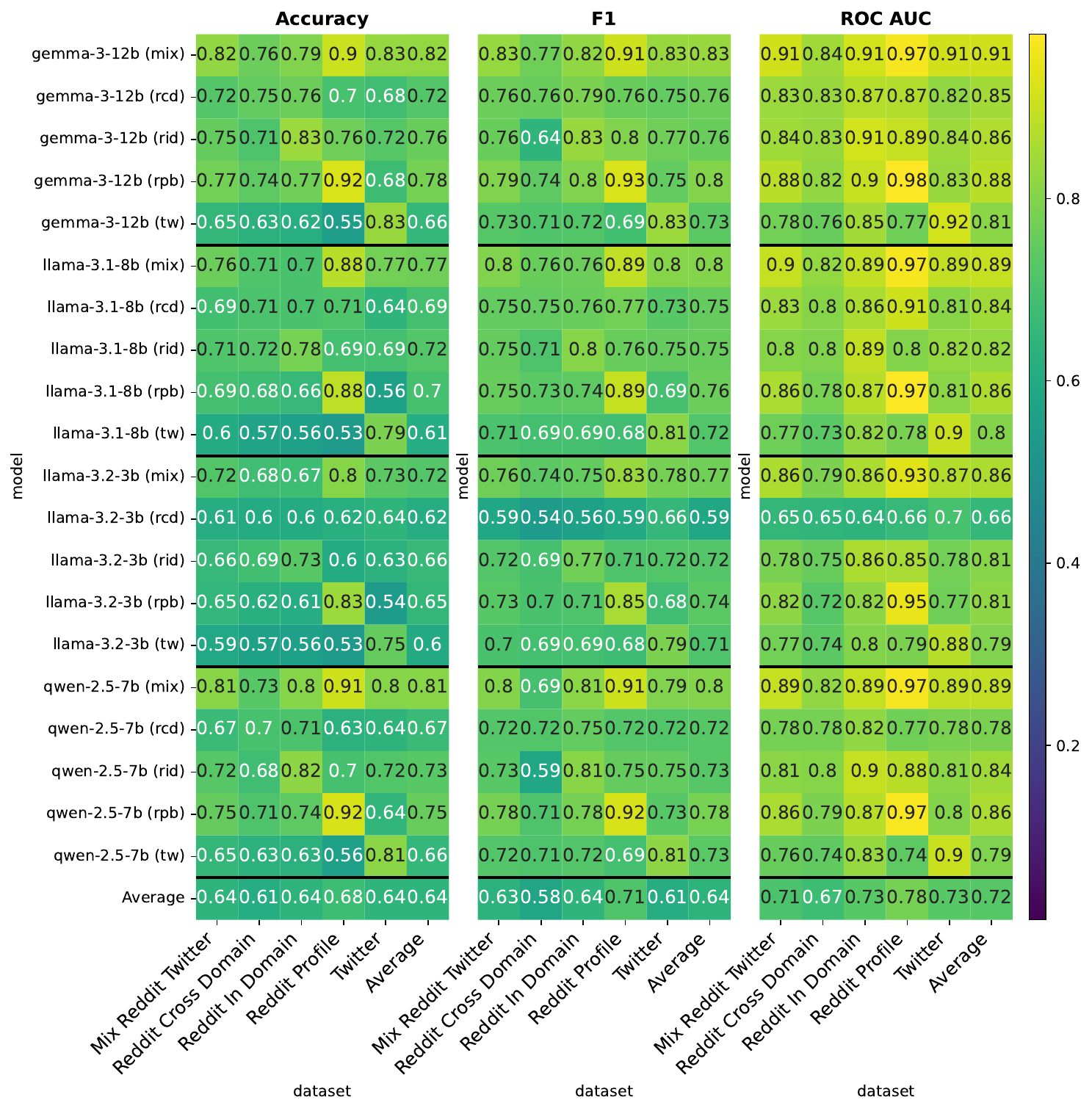}
    \caption{Accuracy, F1-Score and RoC\_AuC Score for our GerAV models. The y-axis shows the model names and the x-axis shows the evaluated test set. The brackets behind the model indicate its training set: \textbf{Mix} Reddit Twitter (mix), \textbf{R}eddit \textbf{C}ross \textbf{D}omain (rcd), \textbf{R}eddit \textbf{I}n \textbf{D}omain (rid), \textbf{R}eddit \textbf{P}rofile \textbf{B}ased (rpb) and \textbf{Tw}itter (tw). The highest value in every column is written in bold.}
    \label{fig:alias_results}
\end{figure*}

Figure \ref{fig:alias_results_baslines} show the results of all baseline model variations (with thresholds tuned on different GerAV subsets), and Figure \ref{fig:alias_results} shows the results of all tuned variants. For some baselines, we only tune the threshold on the mixed validation set, because mixed shows the strongest results for most models. 
The German-pretrained LLäMmlein-7B-Chat model does not outperform the English-pretrained models. Compared to the best-performing model, Gemma-3, it consistently underperforms across fine-tuned variants and evaluation datasets, with an average deficit of 0.04 points F1 in the mixed setting. When compared to Qwen-2.5-7B, which has a the same parameter count, results are more similar: LLäMmlein outperforms Qwen in two of the five fine-tuning configurations (reddit cross-domain and reddit in-domain) on average, while Qwen achieves better results in the remaining three. Overall, these findings suggest that AV fine-tuning does not significantly benefit from German pretraining. This difference may also reflect the overall lower baseline performance of the LLäMmlein model compared to the other models, rather than an effect of its German pretraining, and should be tested on further models and lanugages.
 \FloatBarrier

\FloatBarrier

\section{Word Length buckets}
Table \ref{tab:length_bins} displays the distribution of samples across log-spaced length bins on the mixed test set. Bins with less than 2 samples were excluded, resulting in 27 distinct bins.
\section{Varying Number of Training Samples}
\label{sec:varying_training_number}

\label{sec:bins}
\begin{table}[!ht]
\centering
\begin{tabular}{lrr}
\hline
\textbf{Bin} & \textbf{Range} & \textbf{Samples} \\
\hline
0  & [25.00, 29.96]     & 3604 \\
1  & [29.96, 35.90]     & 3340 \\
2  & [35.90, 43.01]     & 2310 \\
3  & [43.01, 51.54]     & 1053 \\
4  & [51.54, 61.76]     & 896 \\
5  & [61.76, 74.00]     & 784 \\
6  & [74.00, 88.67]     & 583 \\
7  & [88.67, 106.25]    & 516 \\
8  & [106.25, 127.32]   & 413 \\
9  & [127.32, 152.56]   & 356 \\
10 & [152.56, 182.80]   & 348 \\
11 & [182.80, 219.04]   & 295 \\
12 & [219.04, 262.47]   & 210 \\
13 & [262.47, 314.51]   & 190 \\
14 & [314.51, 376.86]   & 145 \\
15 & [376.86, 451.57]   & 113 \\
16 & [451.57, 541.10]   & 97 \\
17 & [541.10, 648.38]   & 73 \\
18 & [648.38, 776.92]   & 55 \\
19 & [776.92, 930.95]   & 41 \\
20 & [930.95, 1115.52]  & 25 \\
21 & [1115.52, 1336.67] & 14 \\
22 & [1336.67, 1601.67] & 14 \\
23 & [1601.67, 1919.21] & 12 \\
24 & [1919.21, 2299.71] & 3 \\
25 & [2299.71, 2755.64] & 10 \\
26 & [2755.64, 3301.96] & 3 \\
27 & [3301.96, 3956.59] & 1 \\
28 & [3956.59, 4741.00] & 1 \\
\hline
\end{tabular}
\caption{Distribution of samples across log-spaced length bins on the mixed test set.}
\label{tab:length_bins}
\end{table}

\begin{figure}[!ht]
    \centering
    \includegraphics[width=1\linewidth]{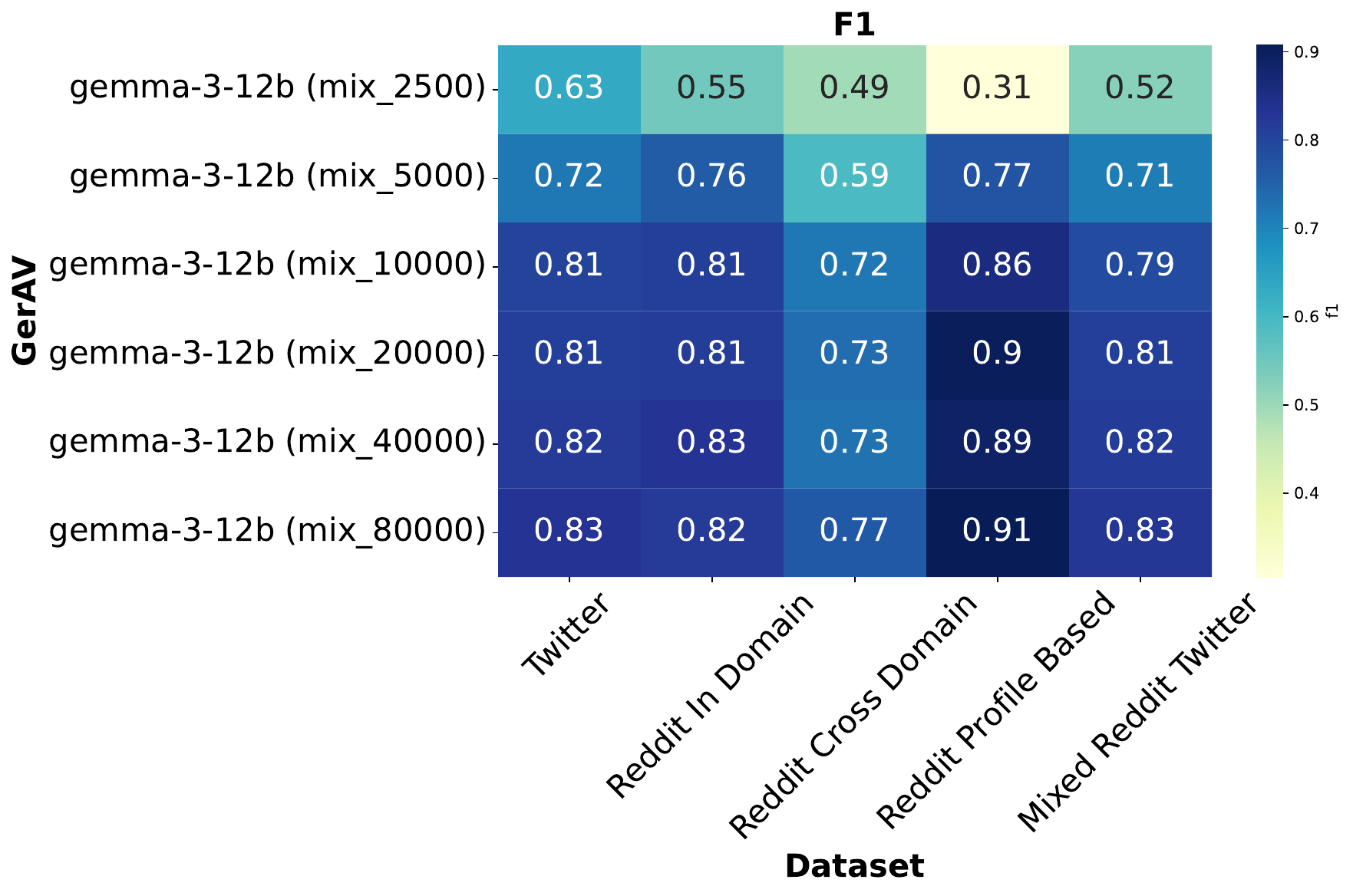}
    \caption{F1-scores for Gemma-3-12b trained on varying numbers of training samples from Mixed Gerav and tested on all GerAV test subsets.}
    \label{fig:no_training_samples}
\end{figure}

Figure \ref{fig:no_training_samples} shows the full results of F1 scores for all training set sizes of GerAV mixed tested on all 5 test subsets. Among these, the Reddit cross-domain test set is most affected by reductions in training data, showing the largest performance decline. This suggests that larger training corpora are particularly important for this challenging setting, likely because increased data diversity helps the model better capture cross-topic stylistic variation required for AV.

\section{Data licence and reproducibility}
\label{sec:data_licence}
Since GerAV is collected from two distinct sources, different access methods and licences apply:
\begin{enumerate}
    \item Twitter: The original Twitter data is publicly accessible with restricted access for academic research via Zenodo\footnote{https://zenodo.org/records/6421331}. The licence of the original data source applies. After gaining access, the GerAV preprocessing and data splits can be reproduced using our code published on GitHub\footnote{https://github.com/NL2G/GerAV/}.
    
    \item Reddit: The Reddit data was collected via the reddit4researchers API \footnote{https://support.reddithelp.com/hc/en-us/articles/14945211791892-Developer-Platform-Accessing-Reddit-Data} that prohibits direct redistribution of the data. To support the reproducibility of GerAV, we provide access to the post URLs for academic research purposes only. Further information about access and the preprocessing pipeline can be found on our GitHub repository. After accessing the Reddit posts, the original Reddit licence applies.
\end{enumerate}